\documentclass[preprint]{elsarticle}

\usepackage{lineno,hyperref}
\usepackage{caption}
\usepackage{subcaption}
\usepackage{multirow}
\modulolinenumbers[5]

\journal{ESWA}

\biboptions{authoryear}

\graphicspath {{Figs/}}

\begin{document}

\begin{frontmatter}

\title{Towards Highly Accurate Coral Texture Images Classification Using Deep Convolutional Neural Networks and Data Augmentation}

\author[ccia]{Anabel G\'omez-R\'ios\corref{mycorrespondingauthor}}
\ead{anabelgrios@decsai.ugr.es}
\author[ccia]{Siham Tabik}
\ead{siham@ugr.es}
\author[ccia]{Juli\'an Luengo}
\ead{julianlm@decsai.ugr.es}
\author[dtu]{ASM Shihavuddin}
\ead{shihav@dtu.dk}
\author[vcu]{Bartosz Krawczyk}
\ead{bkrawczyk@vcu.edu}
\author[ccia]{Francisco Herrera}
\ead{herrera@decsai.ugr.es}

\address[ccia]{Dept. of Computer Science and Artificial Intelligence,
  University of Granada, C/ Periodista Daniel Saucedo Aranda, s/n, 18071, Granada, Spain}
\address[dtu]{Dept. of Applied Mathematics and Computer Science, Technical University of Denmark (DTU), Kgs. Lyngby, Denmark}
\address[vcu]{Dept. of Computer Science, Virginia Commonwealth University, USA}

\cortext[mycorrespondingauthor]{Corresponding author}

\begin{abstract}
The recognition of coral species based on underwater texture images pose a significant
difficulty for machine learning algorithms, due to the three following challenges embedded in the
nature of this data: 1) datasets do not include information about the global structure of the coral;
2) several species of coral have very similar characteristics; and 3) defining the spatial borders
between classes is difficult as many corals tend to appear together in groups. For this reason, the 
classification of coral species has always required an aid from a domain expert. The objective of 
this paper is to develop an accurate classification model for coral texture images. Current datasets
contain a large number of imbalanced classes, while the images are subject to inter-class variation. 
We have analyzed 1) several Convolutional Neural Network (CNN) architectures, 2) data augmentation techniques and 3) transfer 
learning. We have achieved the state-of-the art accuracies using different variations of ResNet on 
the two current coral texture datasets, EILAT and RSMAS.
\end{abstract}

\begin{keyword}
Classification \sep Coral Images \sep Deep Learning \sep Convolutional
Neural Networks \sep Inception \sep ResNet \sep DenseNet.
\end{keyword}

\end{frontmatter}


\section{Introduction}

Coral reefs are complex marine ecosystems typical to the warm and
shallow seas of the tropics. The reefs are created by the slow
accumulation of hard calcium carbonate skeletons that hard coral
species leave behind when they die, waiting for another coral to
live in it and expand the reef. Coral reefs are one of the most
valuable ecosystems in the world as they are extremely biodiverse. 
They support up to two million species and a quarter of all marine
life on Earth~\citep{endangeredSpeciesInternational}. They are also
very important from the human point of view
\citep{ferrario2014effectiveness}. Coral species help to clean the
water and remove nitrogen and carbon, they are a source for medicine
research and economic wealth from fishing and tourism, they are also
a natural barrier for coastal protection against hurricanes and
storms and, since many of them are thousands and even millions
years old, their study helps scientists to understand climatic events
of the past.

The study of the distribution of coral reefs over time can provide important clues about the impact of 
global warming and water pollution levels. According to
\citet{endangeredSpeciesInternational}, we have already lost 19\%
coral reefs areas since the 1950s and, according to the International Union for Conservation of Nature
(IUCN) Red List of Threatened Species
\citep{IUCNredListTable}, in 2017 there were 237 threatened species in the evaluated 40\% 
of the estimated total of species. This is due to the fact that coral reefs do not tolerate temperature 
changes and a quarter of the carbon dioxide emissions in the atmosphere is
absorbed by the ocean, in addition to the water pollution and other
problems caused by humans.

With recent advancements in image acquisition technologies and growing interest in this topic
among the scientific community, 
huge amount of data on coral reefs is being collected. However, it is complicated to keep a record of all
coral species because there are thousands of them and the taxonomy is mutable. This is due to new
discoveries made by scientist or because they may change the order, family or genus of existing species 
as they gather more knowledge about them. In addition, some coral species have different sizes, shapes 
and colors, but other coral species seem to be identical for a human observer. As a consequence, a
successful coral classification has always demanded an expert biologist. If we can automate the
classification by using the amount of coral images that is being collected, we can help scientists to
study more closely that amount of data, making an important step towards automatic knowledge
discovery process. In fact, automatizing 
the classification process of coral images has been addressed in a few number of works.
Most of them \citep{beijbom2012automated,pizarro2008towards,shihavuddin2013image,stokes2009automated}
use machine learning models combined, in some cases, with image enhancement techniques and 
feature extractors. Among these works, only \citet{shihavuddin2013image} use several datasets. 

In recent years, Convolutional Neural Networks (CNNs) have shown outstanding accuracies for image
classification \citep{krizhevsky2012imagenet,ILSVRC15}, especially in the field of Computer Vision.
Currently, their applications branch out to a plethora of diverse fields, where analysis of 
image data is required. In biology, CNNs have been evaluated and compared with machine learning
algorithms for wood classification \citep{AFFONSO2017114}. In coral classification, the use of CNNs 
is challenging due to the variance between images of the same class, the lightning variations due to 
the water column or the fact that some coral species tend to appear together. Besides, CNNs need a 
large dataset to achieve a good performance. In practice, two techniques are used to overcome this
limitation: transfer learning and data augmentation. There are some works that use CNNs for
coral classification \citep{elawady2015sparse,mahmood2016automatic,mahmood2016coral}, but they
use popular CNNs, like VGGnet or LeNet and they only use one dataset to test their models.
Besides, they not use EILAT or RSMAS.

We propose to use more capable CNNs to overcome the limitations of previously applied deep learning 
models. We want to develop a much more accurate model approaching the human expert, facing the
specific problems of coral classification using several datasets. In particular, we have considered 
three of the most promising CNNs, Inception v3 \citep{Szegedy_2016_CVPR}, ResNet \citep{He_2016_CVPR} and
DenseNet \citep{huang2017densely}. Inception is a newer version of GoogleNet, which won the 
ImageNet Large Scale Visual Recognition Competition (ISLVRC)
\citep{ILSVRC15} competition in 2014. ResNet won the same competition in 2015 and DenseNet beat the
results of ResNet in 2016. We have considered two underwater coral datasets, RSMAS 
and EILAT \citep{EILATandRSMAS}, and we have compared our results with the 
most accurate model \citep{shihavuddin2013image}. In these datasets, the images are patches of the corals,
that is, they are small and they show a little part of the coral, a texture, not the entire structure 
of the coral.

The contributions of this work are the following:
\begin{itemize}
	\item Study, explore and analyze the performance of the most promising CNNs in the classification 
    of underwater coral texture images.
	\item Analyze the impact of data augmentation on the performance of the coral texture classification
	model.
	\item Compare our results with the state-of-the-art classical methods which require high human
	supervision and intervention.
\end{itemize}

The rest of the paper is organized as follows. An overview of the three considered CNNs is provided
in Section \ref{sec:models}. The challenges of coral classification and related works are given in
Section \ref{sec:relatedworks}. A description of the coral datasets we have used is provided in Section
\ref{sec:datasets}. The experiments and results are given in Section \ref{sec:results} and the final
conclusions of this study are given in Section \ref{sec:conclusions}.

\section{CNN Classification Models}\label{sec:models}
CNNs have achieved outstanding accuracies in a plethora of contemporary applications, automatizing its
design \citep{FERREIRA2018205}. In fact, since  2012 the prestigious ILSVRC competition \citep{ILSVRC15} 
have been won exclusively by CNN models. The CNN layers capture increasingly
complex features as the depth increases. In recent years, these architectures have evolved by increasing
first the depth of the networks, then the width and finally using lower features obtained from the lower
layers into higher layers. This section provides an overview of the CNNs used in this work. We have
considered three influential CNNs, Inception v3 (Subsection \ref{subsec:inception}), ResNet (Subsection
\ref{subsec:resnet}), and one of the newest, DenseNet, (Subsection \ref{subsec:densenet}). Finally,
we describe the optimization techniques that we have used to overcome the small sizes of the considered
datasets (Subsection \ref{subsec:optimizations}).

\subsection{Inception v3} \label{subsec:inception}
GoogLeNet \citep{szegedy2015going} won the ILSVRC in 2014 and it is based on the repetition
of a module called inception. This module have six convolutions and one max-pooling. Four of these
convolutions use a 1$\times$1 kernel, which is introduced to increase the width and the depth of the
network and to reduce the dimensionality when it is necessary. In this sense, a $1\times1$ convolution is
performed before the other two convolutions in the module, a $3\times3$ and a $5\times5$ convolution. 
After all the computation, the output of the module is calculated as the concatenation of the output of 
the convolutions. This module is repeated 9 times and at the end it uses a dropout layer. In total,
GoogLeNet has 22 learnable layers.

\begin{figure}[]
	\centering
    \includegraphics[width=45mm ]{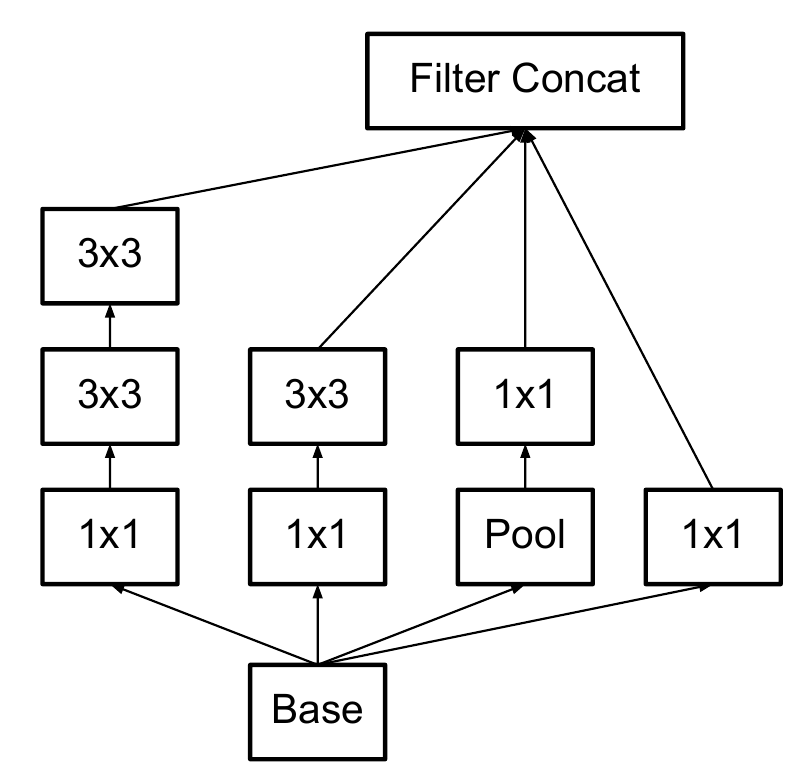}
 \caption{Base Inception v3 module. Figure from \citep{Szegedy_2016_CVPR}.} \label{fig:baseInception}
\end{figure}

Inception v3 can be considered as a modification of GoogLeNet. The base inception module 
is changed by removing the $5\times5$ convolution and introducing instead two $3\times3$ convolutions, 
as we can see in Figure \ref{fig:baseInception}. The resulting network is made up of 10 inception modules. Furthermore, the base module is modified as the network goes deeper. Five modules are changed by
replacing the $n\times n$ convolutions by a $1 \times 7$ followed by a $7 \times 1$ convolution in order 
to reduce the computational cost. The last two modules replace the last two $3 \times 3$ convolutions by a
$1\times3$ and a $3\times 1$ convolutions each one, this time in parallel. Lastly, the first
$7\times7$ convolution in GoogLeNet is also changed by three $3\times 3$ convolutions. In total, 
Inception v3 has 42 learnable layers.

\subsection{ResNet}\label{subsec:resnet}

\begin{figure}[]
	\centering
	\captionsetup[subfloat]{labelformat=empty,justification=centering}
    \includegraphics[width=38mm ]{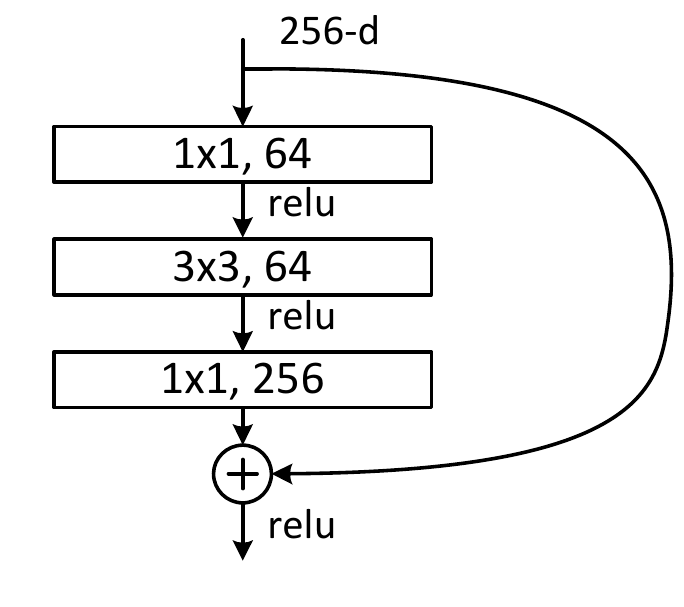}
 \caption{ResNet building block. Figure from \citep{He_2016_CVPR}.} \label{fig:baseResnet}
\end{figure}

Increasing the network depth to obtain a better precision makes the network more difficult to optimize
since it may produce the vanishing or exploding gradients problem. ResNet \citep{He_2016_CVPR}, which 
won the ILSVRC classification task in 2015, address this issue by 
fitting a residual mapping instead of the original mapping, and by adding several connections between
layers. These new connections skip various layers and perform an identity, which 
not adds any new parameters, or a simple $1\times 1$ convolution. In particular, this network is also 
based on the reiterated use of a module, called a building block. The depth of the network depends
on the number of the used building blocks. For 50 or more layers, the building block consists of three
convolutions, a $1\times 1$ followed by a $3\times 3$ followed by a $1\times 1$ convolution, and a
connection joining the input of the first convolution to the output of the third convolution, as we can 
see in Figure \ref{fig:baseResnet}. For our problem, we have used the
model with 50 layers, ResNet-50, and with 152 layers, ResNet-152.

\subsection{DenseNet}\label{subsec:densenet}

\begin{figure}[]
	\centering
	\captionsetup[subfloat]{labelformat=empty,justification=centering}
    \includegraphics[width=60mm ]{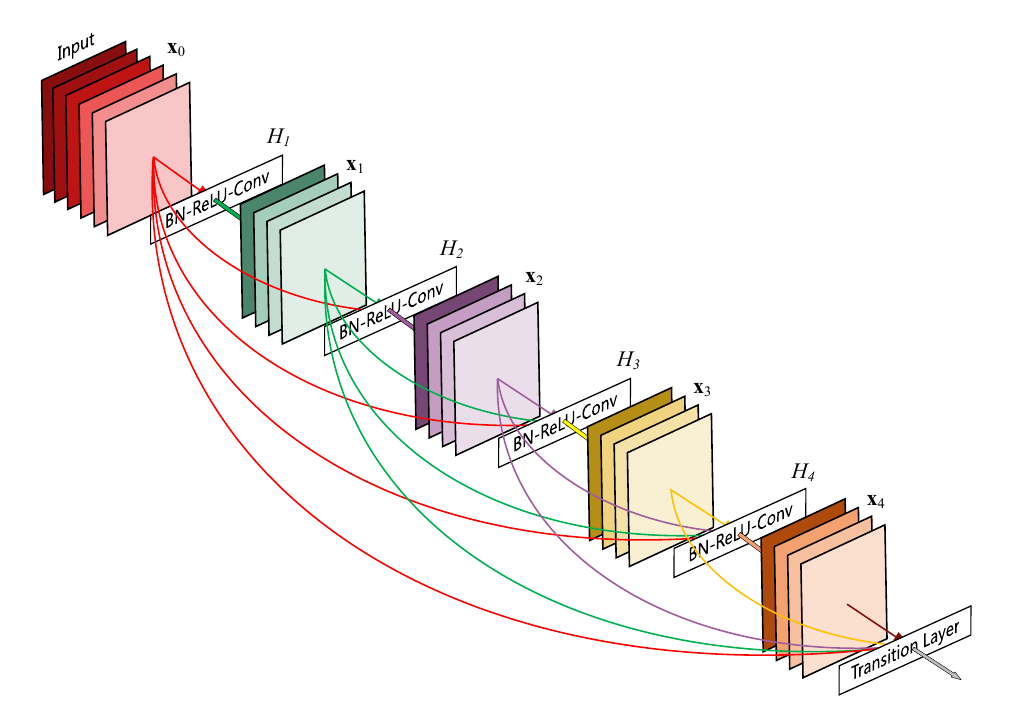}
 \caption{Example of a dense block. Figure from \citep{huang2017densely}.} \label{fig:baseDensenet}
\end{figure}

DenseNet is also based on the repetition of a block, called the dense block. Inspired by the 
building block of ResNet, DenseNet connects the output of all the layers to the 
input of all the following layers within the dense block \citep{huang2017densely}. The connections 
between blocks, called transition layers, work as a compression factor in a sense that the transition
layer generates less feature maps than it receives. The difference between connections
in the dense block and connections in the building block of ResNet is that in the dense block the
outputs of previous layers are added to the following layers before its computation is performed. 
A dense block is the repetition of a Batch Normalization, a ReLU, a $1\times1$ convolution, a Batch
Normalization, a ReLU and a $3\times3$ convolution a specific number of times. In Figure
\ref{fig:baseDensenet} we can see an example of a DenseNet block. The transition layers are a $1\times1$ 
convolution followed by an average pool with kernel $2\times2$. Similarly to ResNet, the number of
dense blocks determines the number of layers in the network. In this work, we have analyze DenseNet-121
and DenseNet-161, which include 121 and 161 layers respectively.

\subsection{CNN Optimization Techniques}\label{subsec:optimizations}
CNN-based models require a large set of training samples to achieve good generalization capabilities.
However, generating large datasets is either costly, time-consuming, or sometimes simply impossible. 
In practice, two techniques are used to overcome this limitation: transfer 
learning and data augmentation. Since the current coral datasets are too small to train an effective
CNN from scratch, we propose to use these two approaches:
\begin{itemize}
	\item \textbf{Transfer learning}: instead of starting the training from a scratch by randomly
	initializing weights, we initialize weights using a pre-trained network on different
	datasets, usually much larger in size. In this work, we have considered using the knowledge
	learned from ImageNet \citep{deng2009imagenet} in our coral classification. We remove the layer 
	that classifies the images into the classes of ImageNet and we add two fully connected layers that
	classify the images into our concrete problem. We train these last two layers, instead of fine-tune
	the whole network, which would also require larger datasets.

	\item \textbf{Data augmentation}: consists of artificially increasing the volume of the
	training set by applying several distortions to the original images, such as changing the 
	brightness, scaling or zooming, rotation, vertical or horizontal mirroring,
	etc. The applied distortions should not alter the spatial pattern of target classes
	\citep{tabik2017snapshot}. Usually the
	distortions are performed during the training time, which allows to do it on the fly without 
	saving the new images.
\end{itemize}

\section{Previous Advances on Automatic Coral Reef Classification}\label{sec:relatedworks}

In this section we explain the challenges of the underwater coral and coral reef images classification
and we give an overview on existing works for automating the classification of coral reef habitat using
underwater imagery. The reasons why the classification of such images is difficult are provided in
Subsection \ref{subsec:challenges}. Previous works in coral classification can be divided into two 
groups, methods that combine classical models (Subsection \ref{subsec:classicalmethods}) and methods 
that use CNNs (Subsection \ref{subsec:cnnmethods}).

\subsection{Challenges of Coral Classification}\label{subsec:challenges}

The classification of underwater coral images is challenging for the following reasons:
\begin{itemize}
	\item Partial occlusion of objects due to the three-dimensional structure of the seabed. Depending on 
    the water type, there can be significant variation in presence of scattering effect, which increases 
    additive noise on the image acquisition and makes it difficult for any computer vision algorithm to 
    perform as in a normal environment.
	\item Lightning variations due to wave focusing and variable optical properties of the water column. 
    In the deep underwater scenario, it is common that there is no natural light source other than the 
    remote sensing device, which implies non uniform illumination across the acquired images.
	\item Subjective annotation of the training samples by different analysts.
    \item Variation in viewpoints, distances and image quality.
	\item Significant inner-class variability in the morphology of benthic organisms.
	\item Complex spatial borders between classes, as many coral species tend to
	appear together. 
    \item The difficulty of keeping the autonomous vehicles stable in the underwater environment,
    which creates significant motion blur when the images are acquired in a video format.
	\item There are very few datasets of underwater coral reef images and in general they contain patches
	of the texture of the coral, while at the same time they do not include any information on the global
	structure of the coral.
\end{itemize}

\subsection{Coral Classification Based on Classical Methods}\label{subsec:classicalmethods}
Most of the existing approaches for classifying underwater coral images
combine one feature extractor with a classifier and show their
performance only using a single dataset, i.e., with specific size,
resolution of the images, number of classes and color information 
\citep{beijbom2012automated,pizarro2008towards,stokes2009automated}. The first paper in this subject
was by \citet{pizarro2008towards}. The authors analyze more marine habitat besides corals, so it is more
general.
They use a SIFT descriptor and a bag of features approach, which means that they choose from the 
training set the images that are more similar to each test image. \citet{beijbom2012automated}
introduced the Moorea Labelled Corals (MLC) dataset, which has large images containing different
coral species, and they used Support Vector Machines along with filters and a texture descriptor. 
They obtained an accuracy of 83.1\% on this dataset using the images of 2008 and 2009 for training and
the images of 2010 for testing. \citet{stokes2009automated} used a normalized color 
space and a discrete cosine transform to extract texture features. Again, they only used one dataset,
provided by the National Oceanic and Atmospheric Administration (NOAA) of the U.S. Department of 
Commerce Ocean Explorer.

The portability of these methods to new datasets has not been
demonstrated yet. Only \citet{shihavuddin2013image} developed an unified classification
algorithm for four different datasets of different characteristics, in which we can find RSMAS
and EILAT. The authors combined multiple image enhancement techniques, feature extractors and 
classifiers, among other techniques. In particular, the image enhancement step contain four algorithms,
one mandatory (Contract Limited Adaptive Histogram Specification or CLAHS) and three optional (color correction, normalization and color channel stretching).
The feature extraction step contain one optional, used as color descriptor, and three mandatory
algorithms, used as texture descriptors. Then, the method has a kernel mapping step with three mandatory
algorithms, a dimension reduction step with two optional algorithms and a prior settings step with one
algorithm. Then it performs the classification using one of the following algorithms: SVM multiclass,
KNN, a neural network or probability density weighted mean distance. Lastly, if the original image was a
mosaic containing several patches, it uses a thematic mapping using sliding window and morphological
filtering. By configuring the hyperparameters and the different combinations of these algorithms, the
model can be adapted to different datasets. 

This method is considered to be the state-of-the-art for RSMAS and EILAT datasets. In particular, in these two datasets
the best combination of algorithms is the following: in the image enhancement step it uses just CLASH.
In the feature extraction step it uses the opponent angle histogram, the hue channel histogram, the
grey level co-occurrence matrix, the completed local binary pattern and the Gabor filter response.
In the kernel mapping step it uses L1 normalization, chi-square kernel and Hellinger kerl for 
the completed local binary pattern and the color histogram. In the dimension reduction step it uses
principal component analysis and Fisher kernel. In the prior settings step it uses class frequency
to estimate prior probability. Finally, as classification algorithm it uses KNN.

As it can be seen, this algorithm implies a lot of human supervision and intervention, 
as there we need to test a lot of algorithms with several hyperparameters and many possible combinations 
between the algorithms. Furthermore, when we have the best combination we need to use a lot of
algorithms every time we need to classify a new image. In the particular case of EILAT and RSMAS, we
need to use six algorithms to obtain the classifier and every time we need to classify a new image
we need to use the first four algorithms, until we obtain the features of the image.

\subsection{Coral Classification Based on CNNs Methods}\label{subsec:cnnmethods}
The use of CNNs for coral classification allow us to use the images without the image enhancements,
although it is possible to use them, and without the feature extraction, saving a lot of experiments to
detect the best combination of algorithms and therefore, saving time.

The first work that used CNNs for coral classification
was by \citet{elawady2015sparse}. The authors first
enhanced the input raw images via color correction and smoothing
filtering. Then, they trained a LeNet-5~\citep{lecun1998gradient} based model
whose input layer consisted of three basic channels of color image
plus extra channels for texture and shape descriptors consisting of
the following components: zero component analysis whitening, phase
congruency, and Weber local descriptor. The model obtained around 55\% accuracy on the two
datasets they used.

\citet{mahmood2016automatic} use VGGnet 
\citep{simonyan2014very} pretrained on ImageNet and the dataset BENTHOZ-2015 \citep{bewley2015australian} 
to fine-tune the network. This dataset contains more than 400,000 images and associated sensor data
collected by an autonomous vehicle over Australia. The authors extract several patches from each image
centered in different pixels and using different scales and they apply a color channel stretch to the
patches as a pre-processing technique. In this article, they propose a mechanism to automatically label
unseen coral reef images to obtain the coral coverage in the region where the images are collected
(i.e., classifying new images as coral or non-coral ones). A marine expert later verifies the accuracy
of this automatic method. In the presented experimental study, authors conducted several experiments
and reported over 90\% accuracy obtained in each of them.

\citet{mahmood2016coral} use the MLC dataset to propose the usage of CNNs along with 
hand-crafted features.
Moreover, they introduce a mechanism to extract such features. This proposal is based on the observation 
that CNNs cannot be trained from scratch using the available coral datasets due to its small size. The
features extraction with CNNs has been carried out with the network VGGnet pre-trained on ImageNet.
To classify both types of features, they use a two layer Perceptron. In their experiments,
they obtain better accuracies with this technique than just with VGGnet or just the hand-crafted
features, although the difference between VGGnet and the combination of the features is small.
In their best experiment, they obtain an accuracy of 84.5\% in MLC.

These works use classical CNNs: VGGnet and LeNet and they do not use EILAT or RSMAS. Besides,
sometimes the accuracies obtained are low \citep{elawady2015sparse}, the classification
is simple \citep{mahmood2016automatic} or they use hand-made feature extraction along with CNNs
\citep{mahmood2016coral}.

 \section{Datasets}\label{sec:datasets}
There exist eight open benchmarks used for underwater coral classification. These include five 
public color datasets: EILAT,
RSMAS, MLC, EILAT2 (a subset of EILAT) and the Red Sea Mosaic dataset. The remaining three are black and
white datasets: UIUCtex, CURET and KTH-TIPS. Some of them include non-coral classes such as fabric, wood 
and brick. It is worth to mention that the Red Sea Mosaic dataset is actually one single large image that
contains a large number of different coral species. 

\begin{figure}
	\centering
	\begin{subfigure}[b][5.4cm][b]{.2\textwidth}
		\centering
		\includegraphics[width=23mm ]{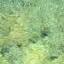}\\\vfill
		\includegraphics[width=23mm ]{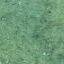}
		\caption{Sand}
	\end{subfigure}
	\begin{subfigure}[b][5.4cm][b]{.2\textwidth}
		\centering
		\includegraphics[width=23mm ]{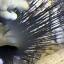}\\\vfill
		\includegraphics[width=23mm ]{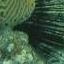}
		\caption{Urchin}
	\end{subfigure}	
	\begin{subfigure}[b][5.4cm][b]{.2\textwidth}
		\centering
		\includegraphics[width=23mm ]{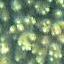}\\\vfill
		\includegraphics[width=23mm ]{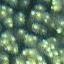}
		\caption{BranchesI}
	\end{subfigure}	
	\begin{subfigure}[b][5.4cm][b]{.2\textwidth}
		\centering
		\includegraphics[width=23mm ]{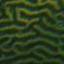}\\\vfill
		\includegraphics[width=23mm ]{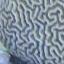}
		\caption{Brain Coral}
	\end{subfigure}
	\begin{subfigure}[b][5.4cm][b]{.2\textwidth}
		\centering
		\includegraphics[width=23mm ]{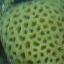}\\\vfill
		\includegraphics[width=23mm ]{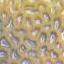}
		\caption{Favid Coral}
	\end{subfigure}	
	\begin{subfigure}[b][5.4cm][b]{.2\textwidth}
		\centering
		\includegraphics[width=23mm ]{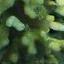}\\\vfill
		\includegraphics[width=23mm ]{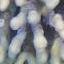}
		\caption{BranchesII}
	\end{subfigure}	
	\begin{subfigure}[b][5.4cm][b]{.2\textwidth}
		\centering
		\includegraphics[width=23mm ]{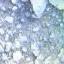}\\\vfill
		\includegraphics[width=23mm ]{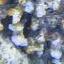}
		\caption{Dead Coral}
	\end{subfigure}
	\begin{subfigure}[b][5.4cm][b]{.2\textwidth}
		\centering
		\includegraphics[width=23mm ]{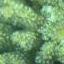}\\\vfill
		\includegraphics[width=23mm ]{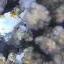}
		\caption{BranchesIII}
	\end{subfigure}
\caption{Selected patches from EILAT. Each column shows two examples per class.} \label{fig:EILAT}
\end{figure}

\begin{figure}
	\centering
	\begin{subfigure}[b][4.4cm][b]{.15\textwidth}
		\centering
		\includegraphics[width=18mm ]{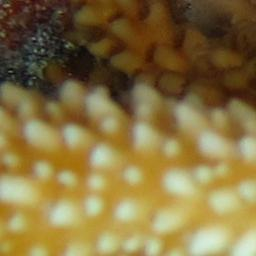}\\\vfill
		\includegraphics[width=18mm ]{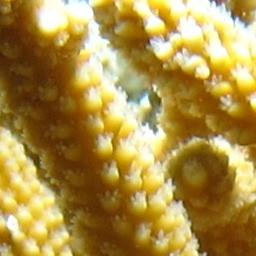}
		\caption{ACER}
	\end{subfigure}
	\begin{subfigure}[b][4.4cm][b]{.15\textwidth}
		\centering
		\includegraphics[width=18mm ]{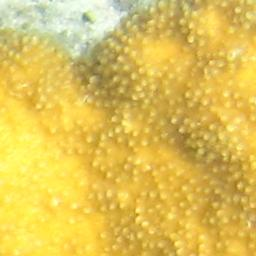}\\\vfill
		\includegraphics[width=18mm ]{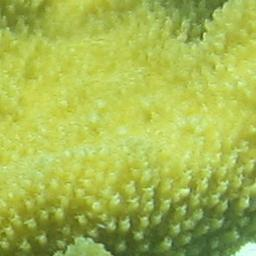}
		\caption{APAL}
	\end{subfigure}	
	\begin{subfigure}[b][4.4cm][b]{.15\textwidth}
		\centering
		\includegraphics[width=18mm ]{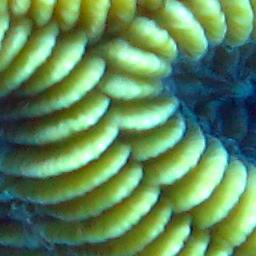}\\\vfill
		\includegraphics[width=18mm ]{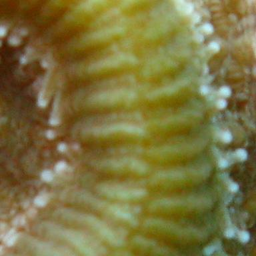}
		\caption{CNAT}
	\end{subfigure}	
	\begin{subfigure}[b][4.4cm][b]{.15\textwidth}
		\centering
		\includegraphics[width=18mm ]{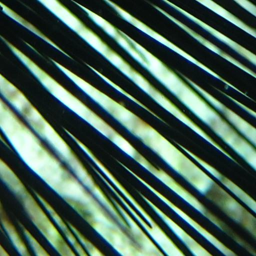}\\\vfill
		\includegraphics[width=18mm ]{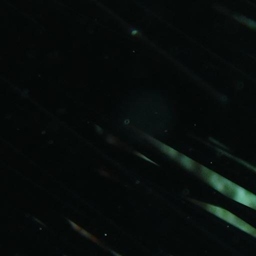}
		\caption{DANT}
	\end{subfigure}
	\begin{subfigure}[b][4.4cm][b]{.15\textwidth}
		\centering
		\includegraphics[width=18mm ]{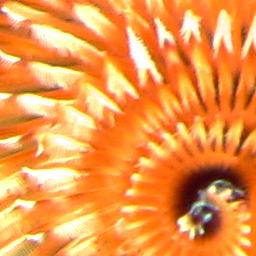}\\\vfill
		\includegraphics[width=18mm ]{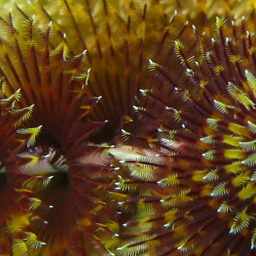}
		\caption{DSTR}
	\end{subfigure}	
	\begin{subfigure}[b][4.4cm][b]{.15\textwidth}
		\centering
		\includegraphics[width=18mm ]{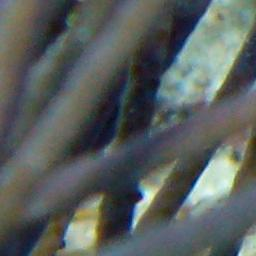}\\\vfill
		\includegraphics[width=18mm ]{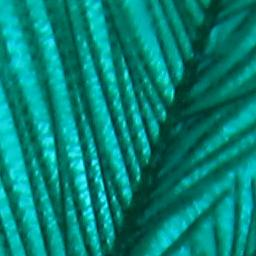}
		\caption{GORG}
	\end{subfigure}	
	\begin{subfigure}[b][4.4cm][b]{.15\textwidth}
		\centering
		\includegraphics[width=18mm ]{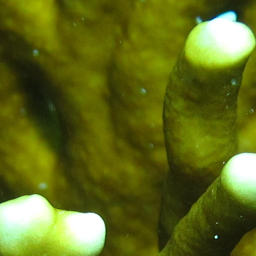}\\\vfill
		\includegraphics[width=18mm ]{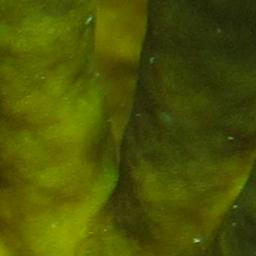}
		\caption{MALC}
	\end{subfigure}
	\begin{subfigure}[b][4.4cm][b]{.15\textwidth}
		\centering
		\includegraphics[width=18mm ]{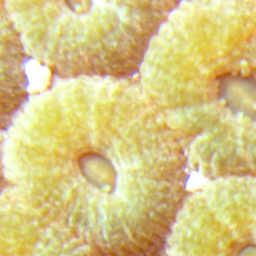}\\\vfill
		\includegraphics[width=18mm ]{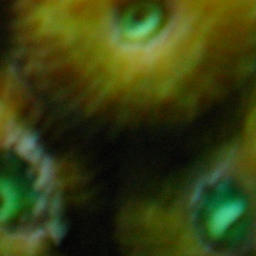}
		\caption{MCAV}
	\end{subfigure}
	\begin{subfigure}[b][4.4cm][b]{.15\textwidth}
		\centering
		\includegraphics[width=18mm ]{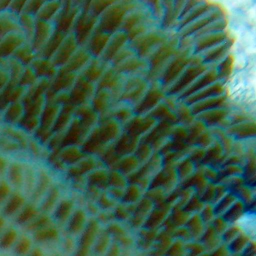}\\\vfill
		\includegraphics[width=18mm ]{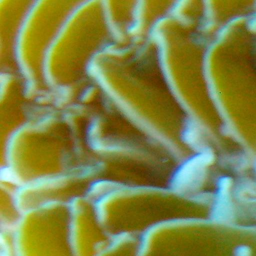}
		\caption{MMEA}
	\end{subfigure}
	\begin{subfigure}[b][4.4cm][b]{.15\textwidth}
		\centering
		\includegraphics[width=18mm ]{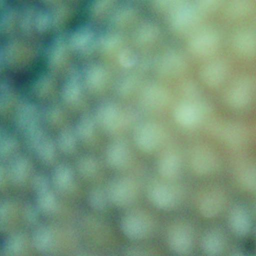}\\\vfill
		\includegraphics[width=18mm ]{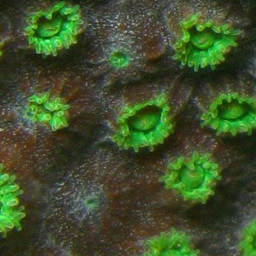}
		\caption{MONT}
	\end{subfigure}
	\begin{subfigure}[b][4.4cm][b]{.15\textwidth}
		\centering
		\includegraphics[width=18mm ]{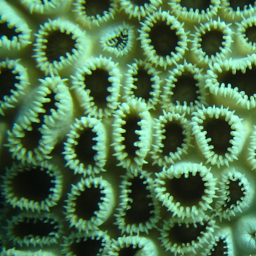}\\\vfill
		\includegraphics[width=18mm ]{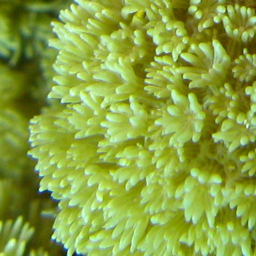}
		\caption{PALY}
	\end{subfigure}
	\begin{subfigure}[b][4.4cm][b]{.15\textwidth}
		\centering
		\includegraphics[width=18mm ]{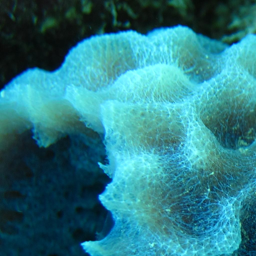}\\\vfill
		\includegraphics[width=18mm ]{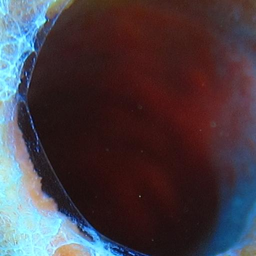}
		\caption{SPO}
	\end{subfigure}
	\begin{subfigure}[b][4.4cm][b]{.15\textwidth}
		\centering
		\includegraphics[width=18mm ]{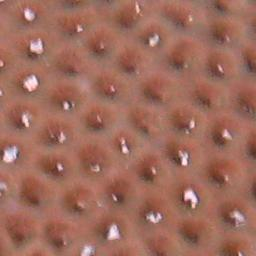}\\\vfill
		\includegraphics[width=18mm ]{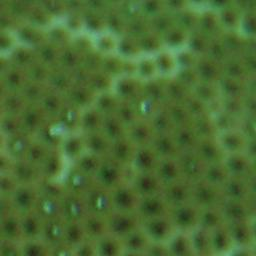}
		\caption{SSID}
	\end{subfigure}
	\begin{subfigure}[b][4.4cm][b]{.15\textwidth}
		\centering
		\includegraphics[width=18mm ]{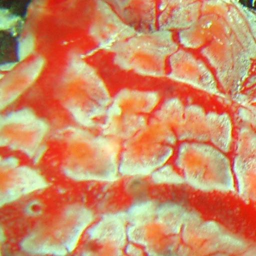}\\\vfill
		\includegraphics[width=18mm ]{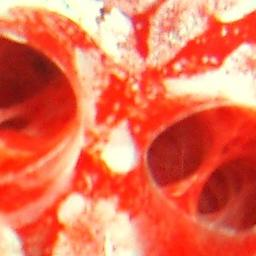}
		\caption{TUNI}
	\end{subfigure}
\caption{Selected patches from RSMAS. Each column shows two examples per class. } \label{fig:RSMAS}
\end{figure}

In this work, we have used the most recent RGB datasets that contain the highest number of corals, 
RSMAS and EILAT \citep{EILATandRSMAS}. These two datasets are comprised of patches of coral images. These 
patches capture mainly the texture of different parts of the coral and do not include any information 
of the global structure of the entire coral. The usage of CNNs with texture images has already been 
successfully carried out for granite tiles classification \citep{FERREIRA20171}.
In this case, both datasets are also highly imbalanced. Some
classes have a high number of samples whereas other classes include very few samples, which makes the
classification more difficult. The main characteristics of these two datasets are as follows:

\begin{itemize}
	\item EILAT contains 1123 image patches of size $64 \times 64$, taken from coral reefs near 
    Eilat in the Red sea. The image patches are pieces of larger images. The original images were
    taken under equal conditions and with the same camera. See examples of patches in Figure
    \ref{fig:EILAT}. The patches have been classified into eight classes, but the used labels do not
    correspond to the coral species names. EILAT is characterized by imbalanced distribution of
    examples among classes, as it can be seen in Table \ref{tab:databases}. 
    
	\item RSMAS contains 766 image patches of size 256$\times$256. The images were collected by divers
	from the Rosenstiel School of Marine and Atmospheric Sciences of the University of Miami. These
	images were taken under different conditions as they were taken with different cameras in 
	different places. See examples in Figure \ref{fig:RSMAS}. The patches have been classified into
	14 classes, whose labels correspond to the names of the coral species in Latin, as it can be seen in
    Table \ref{tab:databases}.
\end{itemize}

\begin{table}[]
\centering 
\footnotesize
\caption{Characteristics of EILAT, RSMAS and StructureRSMAS. The \#imgs refers to the number of images in the corresponding class.} \label{tab:databases}
\begin{tabular}{l|l|l}
 Dataset & Classes & \#imgs \\
 \hline
 EILAT & Sand. & $87$ \\
                & Urchin. & $78$ \\
                & Branches Type I. & $29$\\
                & Brain Coral. & $160$ \\
                & Favid Coral. & $200$ \\
                & Branches Type II. & $216$\\
                & Dead Coral. & $296$\\
                & Branches Type III. & $11$ \\
\hline
RSMAS & Acropora Cervicornis (ACER). & $109$\\
                & Acropora Palmata (APAL). & $77$ \\
                & Colpophyllia Natans (CNAT). & $57$ \\
                & Diadema Antillarum (DANT). & $63$ \\
                & Diploria Strigosa (DSTR). & $24$ \\
                & Gorgonians (GORG). & $60$ \\
                & Millepora Alcicornis (MALC). & $22$ \\
                & Montastraea Cavernosa (MCAV). & $79$ \\
                & Meandrina Meandrites (MMEA). & $54$ \\
                & Montipora spp. (MONT). & $28$ \\
                & Palythoas Palythoa (PALY). & $32$ \\
                & Sponge Fungus (SPO). & $88$ \\
                & Siderastrea Siderea (SSID). & $37$ \\
                & Tunicates (TUNI). & $36$ \\
\end{tabular}
\end{table}

\section{Classification of Coral Texture Images with CNNs}\label{sec:results}

This section is organized in four parts. First, we describe the experimental framework we have used 
(Subsection \ref{subsec:framework}). Second, we analyze the results of our CNN-based classifiers without
data augmentation and compare them with the state-of-the-art classical models on EILAT and RSMAS
(Subsection \ref{subsec:texture}). Third, we analyze the impact of data augmentation on these two
texture datasets (Subsection \ref{subsec:dataaugmentation}). Finally, we provide a deeper 
analysis on the missclassified EILAT and RSMAS images by their best models (Subsection
\ref{subsec:missclassified}). 

\subsection{Experimental Framework}\label{subsec:framework}

All the results provided in this section have been obtained performing a 5 fold cross validation 
technique. To analyze and compare the performance of different CNNs architectures, configurations and 
optimizations, we have used the mean of the five accuracies obtained in the five folds. The
accuracy is calculated as follows:

$$ \mathrm{Accuracy} = \frac{\mathrm{True\ Positives} + \mathrm{True\ Negatives}}{N} \enspace , $$

where $N$ is the total number of instances.

To evaluate ResNet and DenseNet, we have used Keras \citep{chollet2015keras} as front-end and Tensorflow
\citep{tensorflow2015-whitepaper} as back-end. To evaluate
Inception, we have used its implementation in Tensorflow. We have used transfer learning by initializing
the considered CNNs with the pre-trained weights of the networks on Imagenet. We have also analyzed 
the impact of these data augmentation techniques on the performance of the learning process:
\begin{itemize}
	\item Random shift (referred to later as shift) consists of randomly shifting the 
    images horizontally or vertically by a factor calculated as the fraction of the width or 
    length of the image. In this work we shift the images horizontally and vertically in all the
    cases. Given a number $x$, the width and length of the image will be shifted by a random factor
    selected in the interval $[0,x]$. 
    \item Random zoom (referred to later as zoom) consists of randomly zooming the image by
    a certain range. Given a value $x$, each image will be resized in the interval $[1-x, 1+x]$.
    \item Random rotation (referred to later as rotation) consist of randomly rotating the
    images by a certain angle. Given a value $x$, each image will be rotated by an angle in $[0,x]$.
    \item Random horizontal flip (referred to later as flip) consist of randomly flipping 
    the images horizontally.
\end{itemize}

\begin{figure}
	\centering
	\begin{subfigure}[b][3.8cm][b]{.3\textwidth}
		\centering
		\includegraphics[width=30mm ]{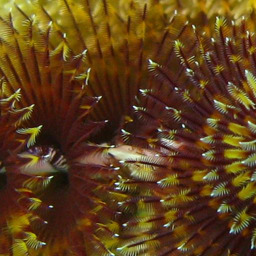}
		\caption{Original}\label{fig:dataAugOriginal}
	\end{subfigure}
	\begin{subfigure}[b][3.8cm][b]{.3\textwidth}
		\centering
		\includegraphics[width=30mm ]{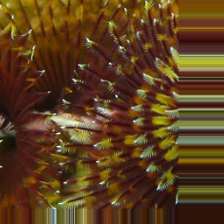}
		\caption{Shift}\label{fig:dataAugShift}
	\end{subfigure}
	\begin{subfigure}[b][3.8cm][b]{.3\textwidth}
		\centering
		\includegraphics[width=30mm ]{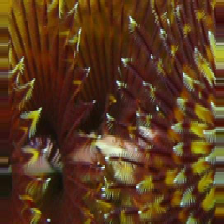}
		\caption{Zoom}\label{fig:dataAugZoom}
	\end{subfigure}	
	\begin{subfigure}[b][3.8cm][b]{.3\textwidth}
		\centering
		\includegraphics[width=30mm ]{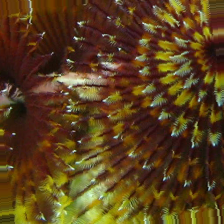}
		\caption{Rotation}\label{fig:dataAugRotation}
	\end{subfigure}
	\begin{subfigure}[b][3.8cm][b]{.3\textwidth}
		\centering
		\includegraphics[width=30mm ]{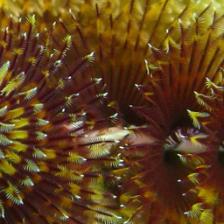}
		\caption{Flip}\label{fig:dataAugFlip}
	\end{subfigure}	
\caption{The result of applying four data augmentation techiques to (\textbf{a}) a original RSMAS image: (\textbf{b}) shift, (\textbf{c}) zoom, (\textbf{d}) rotation and (\textbf{e}) flip.} \label{fig:dataAug}
\end{figure}	

An illustration of these data augmentation techniques is shown in Figure \ref{fig:dataAug}. As it can be seen, the 
distorted images maintain the original size and the pixels outside the boundaries are filled with the
values of the limit pixels. This effect can be clearly seen in Figure \ref{fig:dataAugShift}. 

Besides these data augmentation techniques, we have also evaluated the impact of different
hyperparameters on the performance of the analyzed networks, such as the number of iterations, the 
batch size and the number of layers.

\subsection{Classification of Coral Texture Images without Data Augmentation}\label{subsec:texture}

\begin{table}[]
\centering
\caption{The accuracies obtained by Inception v3, ResNet-50, ResNet-152, DenseNet-121, DenseNet-161 and the classical state-of-the-art Shihavuddin model. The results of all the Convolutional Neural Networks (CNNs) were obtained without data augmentation. The best results are stressed in bold.}
\label{tab:results}
\resizebox{\textwidth}{!}{\begin{tabular}{l|llllll}
      & \begin{tabular}[c]{@{}l@{}}Shihavuddin's\\method\end{tabular} & Inception v3 & ResNet-50 & ResNet-152  & DenseNet-121 & DenseNet-161 \\ \hline
EILAT & 95.79      & 95.25    & \textbf{97.85}    &  \textbf{97.85}     & 91.03       & 93.81       \\
RSMAS & 92.74      & 96.03    & 97.67    &  \textbf{97.95}     & 89.73       & 91.10
\end{tabular}}
\end{table}

\begin{table}[]
\centering
\caption{The set of hyperparameters that provides the best performance shown in Table \ref{tab:results} for each CNN model.}
\label{tab:parameters}
\resizebox{\textwidth}{!}{\begin{tabular}{ll|lllll}
                                            &            & Inception v3 & ResNet-50 & ResNet-152 & DenseNet-121 & DenseNet-161 \\ \hline
\multicolumn{1}{l|}{\multirow{2}{*}{EILAT}} & Batch Size & 100          & 64        & 64         & 32           & 32           \\
\multicolumn{1}{l|}{}                       & Epochs     & 4000         & 500       & 300        & 300          & 700          \\ \hline
\multicolumn{1}{l|}{\multirow{2}{*}{RSMAS}} & Batch Size & 100          & 64        & 32         & 32           & 64           \\
\multicolumn{1}{l|}{}                       & Epochs     & 4000         & 1300      & 300        & 700          & 1000        
\end{tabular}}
\end{table}

In this subsection we have evaluated exhaustively Inception v3, ResNet and DenseNet with different
hyperparameters and we have compared the results obtained for these three CNNs with the state-of-the-art
model by \citet{shihavuddin2013image} on EILAT and RSMAS. For Inception, 
we have analyzed the impact of different numbers of iterations and batch sizes. For ResNet and DenseNet,
we have evaluated different combinations of number of epochs, batches sizes and network depths.

As the results provided by \citet{shihavuddin2013image} were performed using a 10 fold
cross validation, we had to re-evaluate their model using a 5 fold cross validation with the same
folds we have used for all other models in order to compare them under the same conditions. 
We have also used the best hyperparameters for each dataset at each step of the method described
in Subsection \ref{subsec:classicalmethods}.

The results of Shihavuddin's method, Inception v3, ResNet with 50 and 152 layers and DenseNet with 121 
and 161 layers are shown in Table \ref{tab:results}, while the corresponding best configurations are 
shown in Table \ref{tab:parameters}. As it can be seen from this table, ResNet-152 outperforms
Shihavuddin's model and the rest of the CNN models. Inception provides a better accuracy than 
Shihavuddin's method on RSMAS, but shows a worse accuracy on EILAT. DenseNet shows the worst results
on both datasets. In general, these results show that CNNs are able to become the 
state-of-the-art in coral classification tasks. In RSMAS, the best model is ResNet-152, with more than a 
5\% improvement with respect to Shihavuddin's method. In EILAT, ResNet-50 and ResNet-152 achieve
exactly the same accuracy, it is not a cause of rounding, and they outperform Shihavuddin's method for more than 2\%.

Obtained results allow us to conclude that only by training the last layers of a CNN that is
already pre-trained on ImageNet and without data augmentation, which is the technique that usually 
takes more time, we can outperform a method that takes long running times and need high human
supervision. In fact, Shihavuddin's method is composed of six steps and each step is composed of one or
various algorithms. Then, in order to obtain the best performance, it is needed to evaluate all the
possible algorithm combinations through all the steps and to optimize the hyperparameters of each
algorithm. Furthermore, this has to be done independently for each dataset we want to classify.

\subsection{Classification of Coral Texture Images with Data Augmentation}\label{subsec:dataaugmentation}
In this subsection we have analyzed the effect of the data augmentation techniques listed in
Subsection \ref{subsec:framework} on the classification of texture images. To carry out this analysis,
we have selected the best performing model for each dataset. In EILAT, ResNet-50 and ResNet-152 are 
the best models and provide the same accuracy, so we have chosen ResNet-50 as it is simpler and has 
less parameters. In RSMAS, the best model is ResNet-152.

Recall that if we note \texttt{rotation = 2}, it means that we are applying a random rotation to the
images by an angle in the interval $[0,2]$. This notation is equivalent for all the other techniques.

\begin{table}[]
\centering
\caption{The accuracies obtained by the best performing CNN on EILAT, ResNet-50, with different data augmentation techniques using the set of hyperparameters indicated in Table \ref{tab:parameters}. The best result is stressed in bold.}
\label{tab:dataAugEILAT}
\resizebox{\textwidth}{!}{\begin{tabular}{l|llllll}
         & \begin{tabular}[c]{@{}l@{}}without data\\ augmentation\end{tabular} & shift = 0.2 & zoom = 0.2 & rotation = 2 & flip   & \begin{tabular}[c]{@{}l@{}}shift = 0.2,\\ zoom = 0.2\end{tabular} \\ \hline
Accuracy & 97.85 & \textbf{98.03}      & 97.85     & 97.40       & 97.53 & 97.85                 
\end{tabular}}
\end{table}

\begin{table}[]
\centering
\caption{The accuracies obtained by the best performing CNN on RSMAS, ResNet-152, with different data augmentation techniques  using the set of hyperparameters indicated in Table \ref{tab:parameters}. The best result is stressed in bold.}
\label{tab:dataAugRSMAS}
\resizebox{\textwidth}{!}{\begin{tabular}{l|llllll}
         & \begin{tabular}[c]{@{}l@{}}without data\\ augmentation\end{tabular} & shift = 0.2 & zoom = 0.4 & rotation = 2 & flip   & \begin{tabular}[c]{@{}l@{}}shift = 0.2,\\ zoom = 0.4\end{tabular} \\ \hline
Accuracy & 97.95 & 98.36      & \textbf{98.63}     & 97.40       & 97.578 & 98.08                                                           
\end{tabular}}
\end{table}

Tables \ref{tab:dataAugEILAT} and \ref{tab:dataAugRSMAS} show respectively the results of ResNet-50
and ResNet-152 on EILAT and RSMAS using data augmentation together with the parameters that provide
the best performance. The number of steps is the number of times that we generate a batch of new images
by data augmentation at each epoch. As the number of steps increases, the accuracy improves but also the
time needed to complete each experiment increases. In this case, 300 was the best number of steps.
Although we are only showing the best accuracies we have obtained 
from using data augmentation, the difference in accuracy between the best data augmentation technique 
and the results obtained without data augmentation is quite small. In both cases the improvement is 
less than 1\%. 

This slight improvement using data augmentation can be explained by the nature of used images. Since the
original images are small and close-up, the applied modifications do not have much effect on the
learning of the models as they need to be small: the shift implies to loose part of the images, and they
are already very small; and the zoom implies to loose quality of the images, and they are already blurry
because they are underwater images. On the other hand, the images are so close-up that the rotation and
the flipping do not introduce significant variations among them. Besides, the performance of the base
models is already good without any data augmentation. Therefore, we can conclude that the use of data
augmentation techniques in texture coral images does not significantly improve the learning model.

\subsection{Analyzing the Missclassified Images}\label{subsec:missclassified}
In this subsection we have analyzed the missclassified images in each partition of the 5 fold cross
validation in both datasets, EILAT and RSMAS.

\begin{figure}
	\centering
	\begin{subfigure}[b][2.7cm][b]{\textwidth}
		\centering
		\includegraphics[width=18mm ]{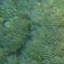}
		\includegraphics[width=18mm ]{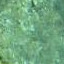}
		\includegraphics[width=18mm ]{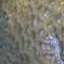}
		\includegraphics[width=18mm ]{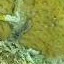}
		\includegraphics[width=18mm ]{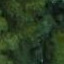}
		\caption{Examples of five images missclasified as Dead Coral}
	\end{subfigure}
	\begin{subfigure}[b][2.7cm][b]{\textwidth}
		\centering
		\includegraphics[width=18mm ]{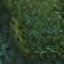}
		\includegraphics[width=18mm ]{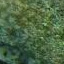}
		\includegraphics[width=18mm ]{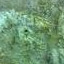}
		\caption{Dead Coral images}
	\end{subfigure}	
\caption{Examples of (\textbf{a}) missclassified images in EILAT as Dead Coral and (\textbf{b}) original Dead Coral images.} \label{fig:badEILAT}
\end{figure}

\begin{figure}
	\centering
	\begin{subfigure}[b][2.7cm][b]{\textwidth}
		\centering
		\includegraphics[width=18mm ]{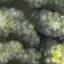}
		\includegraphics[width=18mm ]{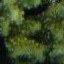}
		\includegraphics[width=18mm ]{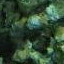}
		\caption{Branches Type III images}
	\end{subfigure}
	\begin{subfigure}[b][2.7cm][b]{\textwidth}
		\centering
		\includegraphics[width=18mm ]{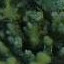}
		\includegraphics[width=18mm ]{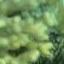}
		\includegraphics[width=18mm ]{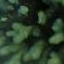}
		\includegraphics[width=18mm ]{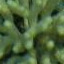}
		\caption{Branches Type II images}
	\end{subfigure}	
\caption{Examples that show the similarities between (\textbf{a}) Branches Type III and (\textbf{b}) Branches Type II. The third image form (\textbf{a}) is missclasified as Branches Type II. The first and second images from (\textbf{b}) are missclassified as Branches Type III.} \label{fig:badEILATtwo}
\end{figure}

In EILAT, ResNet-50 produced 22 missclassified images in all of the test folds. 14 of this missclassified 
images have been classified as Dead Coral. Dead Coral is the class with the highest number of images as
all the dead corals (no matter what species) are in this class. This implies that this class shares some
features with all the other classes, as we can see in Figure \ref{fig:badEILAT}. Additionally, this
may also be caused by a classification bias emerging due to the imbalance nature of EILAT
\citep{Krawczyk2016}. Similarly, there
are four Dead Coral images classified as other classes. In total, 18 of 22 images are 
missclassified due to this class. The remaining three images are from the class
Branches Type II missclassified as Branches Type III or vice versa. As we can observe in Figure
\ref{fig:badEILATtwo}, some images in these two classes are very similar and therefore it is very
difficult to distinguish between them. 

\begin{figure}
	\centering
	\begin{subfigure}[b][2.7cm][b]{\textwidth}
		\centering
		\includegraphics[width=18mm ]{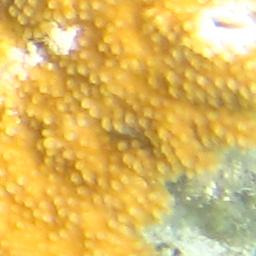}
		\includegraphics[width=18mm ]{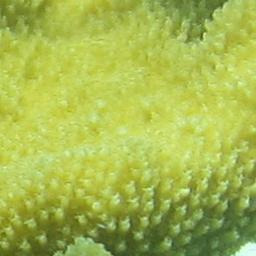}
		\includegraphics[width=18mm ]{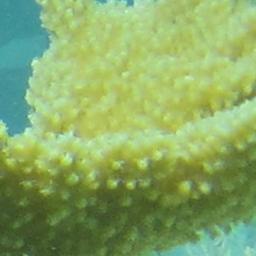}
		\includegraphics[width=18mm ]{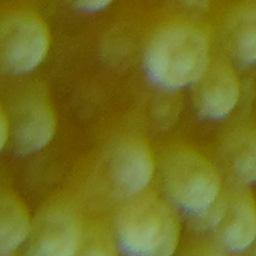}
		\caption{APAL}
	\end{subfigure}
	\begin{subfigure}[b][2.7cm][b]{\textwidth}
		\centering
		\includegraphics[width=18mm ]{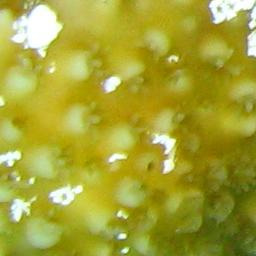}
		\includegraphics[width=18mm ]{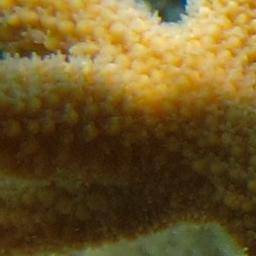}
		\includegraphics[width=18mm ]{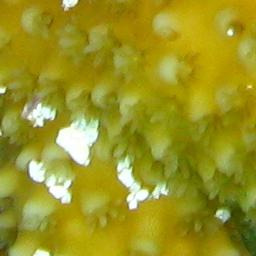}
		\includegraphics[width=18mm ]{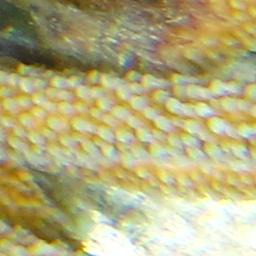}
		\caption{ACER}
	\end{subfigure}	
\caption{Examples that show the similarities between (\textbf{a}) APAL and (\textbf{b}) ACER. The third and fourth images from (\textbf{a}) are missclassified as ACER. The first and second images from (\textbf{b}) are missclassified as APAL.} \label{fig:badRSMAS}
\end{figure}	

\begin{figure}
	\centering
	\begin{subfigure}[b][2.7cm][b]{\textwidth}
		\centering
		\includegraphics[width=18mm ]{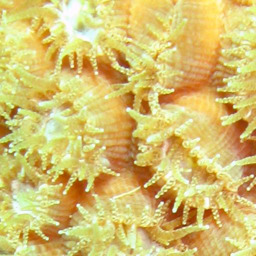}
		\includegraphics[width=18mm ]{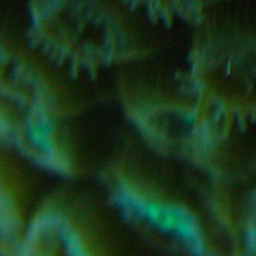}
		\includegraphics[width=18mm ]{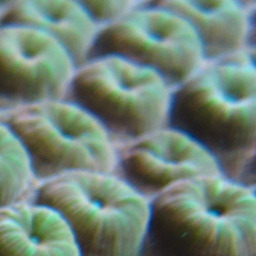}
		\caption{MCAV}
	\end{subfigure}
	\begin{subfigure}[b][2.7cm][b]{\textwidth}
		\centering
		\includegraphics[width=18mm ]{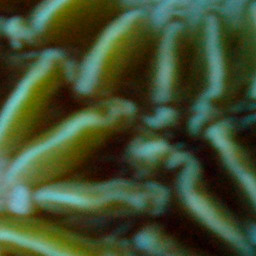}
		\includegraphics[width=18mm ]{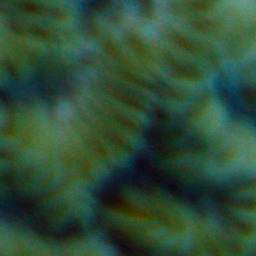}
		\includegraphics[width=18mm ]{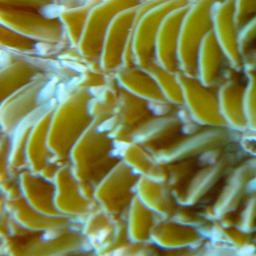}
		\caption{MMEA}
	\end{subfigure}	
\caption{Examples that show the similarities between (\textbf{a}) MCAV and (\textbf{b}) MMEA. The second and third images in (\textbf{a}) are missclassified as MMEA.} \label{fig:badRSMAStwo}
\end{figure}	

In RSMAS, ResNet-152 produced only 10 missclassified images in all of the test folds. In general, the model
tends to missclassify APAL as ACER and vice versa. The model always missclassified MCAV as
MMEA. We can see the similarities between these classes
in Figures \ref{fig:badRSMAS} and \ref{fig:badRSMAStwo}. The rest of the missclassified images are blurry
images.

From these missclassified images, we can conclude that in the case of EILAT it would be needed an
expert to distinguish between the images in the class Dead Coral and the rest of the classes, as
the images are very similar. For the images in Branches Type II and Type III a good solution
might be for an expert to reclassify the images into more specific classes, like the coral species,
which we have seen with RSMAS that is a good option. In the case of RSMAS, the missclassified 
images are again between classes that looks very similar between them, so we would need an expert
to distinguish between them. In this case, as the images are so close-up, maybe it would be a
good solution to make use of images from the same species that contain the whole coral body.

\section{Conclusions}\label{sec:conclusions}

The classification of underwater coral images is challenging due to the large number of different coral
species, the great variance among images of the same coral species, the lightning variations due to the
water column, or the fact that several species tend to appear together, leading to an increasing
overlapping among different classes. Few works have tackle this problem, but the only one that 
classifies EILAT and RSMAS is a really complex method which makes use of several algorithms and takes
a lot of human intervention and time. We have addressed these problems by using some of the most 
powerful CNNs, namely Inception v3, ResNet and DenseNet. We have carried out a study of the foundations
of this three CNNs, their parameter set-up, and possibility of using data augmentation techniques
to aid their learning process. We have been able to outperform the state-of-the-art approach, proving
that CNNs are an excellent technique for automatic classification of underwater coral images.

We have showed that CNNs based models achieved the state-of-the-art accuracies on the coral datasets
RSMAS and EILAT, surpassing classical methods that require a high human intervention, and without using
data augmentation. In particular, ResNet have been the best CNN in RSMAS and EILAT. 

When considering the impact of data augmentation, we have shown that from these two datasets, which
contain very close-up images taken under similar conditions and have a lot of inner-class variance,
there is a little benefit obtained from using such techniques.

This work enables new advanced challenges like classifying not just texture coral images, but structure
coral images too. In particular, the problem of classifying any coral image using a single classifier,
either texture or structure, will be addressed.

\section*{Acknowledgments}
This work was partially supported by the Spanish Ministry of Science and
Technology under the project TIN2017-89517-P and by the
Andalusian Government under the project P11-TIC-7765. Siham Tabik was
supported by the Ramón y Cajal Programme (RYC-2015-18136) and Anabel G\'omez-R\'ios
was supported by the FPU Programme 998758-2016. The Titan X Pascal used for this research was donated by the NVIDIA Corporation.


\bibliography{biblio}

\end{document}